\documentclass[preprint,12pt]{elsarticle}
\journal{Neural Networks}

\usepackage{amsthm}
\usepackage{amsmath,amsfonts}
\usepackage{algorithmic}
\usepackage{algorithm}
\usepackage{array}
\usepackage[caption=false,font=normalsize,labelfont=sf,textfont=sf]{subfig}
\usepackage{textcomp}
\usepackage{stfloats}
\usepackage{url}
\usepackage{verbatim}
\usepackage{graphicx}
\usepackage[colorlinks=true, allcolors=blue]{hyperref}
\usepackage{booktabs}
\usepackage{svg}
\usepackage{tikz}
\usetikzlibrary{positioning}
\usepackage{adjustbox}
\usepackage{dcolumn} 
\newtheorem{theorem}{\bf Theorem}
\newtheorem{lemma}{\bf Lemma}
\newtheorem{definition}{\bf Definition}
\usepackage{hyperref}
\usepackage{epstopdf} 

\begin{document}

\begin{frontmatter}

\title{A Differential Manifold Perspective and Universality Analysis of Continuous Attractors in Artificial Neural Networks} 

\author[1]{Shaoxin Tian}
\author[1]{Hongkai Liu}
\author[2]{Yuying Yang}
\author[2]{Jiali Yu\corref{cor1}}
\author[2]{Zizheng Miao}
\author[2]{Xuming Huang}
\author[2]{Zhishuai Liu}
\author[3]{Zhang Yi}

\affiliation[1]{organization={Yingcai Honors College, University of Electronic Science and Technology of China},
            addressline={No.2006, Xiyuan Ave, West Hi-Tech Zone}, 
            city={Chengdu},
            postcode={611731}, 
            state={Sichuan},
            country={China}}
\affiliation[2]{organization={School of Mathematical Sciences, University of Electronic Science and Technology of China},
            addressline={No.2006, Xiyuan Ave, West Hi-Tech Zone}, 
            city={Chengdu},
            postcode={611731}, 
            state={Sichuan},
            country={China}}
\affiliation[3]{organization={College of Computer Sciences, Sichuan University},
            addressline={24 South Secotion 1, 1st Ring Road}, 
            city={Chengdu},
            postcode={610065}, 
            state={Sichuan},
            country={China}}
\cortext[cor1]{Corresponding author: Jiali Yu, email: yujiali@uestc.edu.cn}

\begin{abstract}

Continuous attractors are critical for information processing in both biological and artificial neural systems, with implications for spatial navigation, memory, and deep learning optimization. However, existing research lacks a unified framework to analyze their properties across diverse dynamical systems, limiting cross-architectural generalizability.

This study establishes a novel framework from the perspective of differential manifolds to investigate continuous attractors in artificial neural networks. It verifies compatibility with prior conclusions, elucidates links between continuous attractor phenomena and eigenvalues of the local Jacobian matrix, and demonstrates the universality of singular value stratification in common classification models and datasets. These findings suggest continuous attractors may be ubiquitous in general neural networks, highlighting the need for a general theory, with the proposed framework offering a promising foundation given the close mathematical connection between eigenvalues and singular values.
\end{abstract}

\begin{keyword}
Continuous attractors \sep differentiable manifold \sep eigenvalue \sep point attractors \sep Deep Neural Network \sep singular value

\end{keyword}

\end{frontmatter}



\section{Introduction}\label{sec1}
In recent years, significant advancements have been made in both neuroscience and the intersecting fields of artificial intelligence (AI) and deep learning. In neuroscience, continuous technological innovations have facilitated deeper exploration of brain organization at both microscopic and macroscopic levels. From microscale synaptic connectivity to macroscale interregional coordination, researchers have progressively uncovered the intricate mechanisms underlying information processing, cognition, memory, and decision making \cite{bib1, bib2, bib3, bib4}. A key discovery highlights the essential role of grid cells in spatial navigation \cite{bib1}. These neurons, located in the medial entorhinal cortex of rodents, exhibit periodic hexagonal firing patterns, with their intrinsic dynamics closely associated with low-dimensional continuous attractors. The collective activity of grid cells is constrained within a two-dimensional manifold that remains stable across different environmental conditions. This discovery provides strong evidence that the brain utilizes continuous attractor dynamics for spatial computation, offering crucial insights into the neural mechanisms underlying spatial perception and navigation.

Attractors play a fundamental role in dynamical systems theory, representing stable equilibrium states toward which a system naturally evolves \cite{bib5, bib6, bib7}. A growing body of neuroscientific research suggests that the brain leverages attractor dynamics for information storage, where memories are encoded as stable states within the neural state space \cite{bib2, bib3, bib8}. Upon receiving specific external inputs, neural networks spontaneously transition to the corresponding attractor states, thus activating distinct neuronal firing patterns that facilitate information retrieval \cite{bib6, bib7, bib9}. This principle is well exemplified in memory formation, which can be understood as a process in which various information patterns are progressively shaped into attractors through synaptic plasticity mechanisms \cite{bib2, bib4, bib6}.

Three principal attractor types are distinguished by their topological configurations\cite{bib5, bib14, bib15, bib16, bib17}: Discrete attractors correspond to isolated fixed points encoding categorical information, continuous attractors form manifolds supporting analog variable representation, and periodic attractors generate cyclic state transitions underlying rhythmic neural processes. This classification framework bridges mathematical abstractions with neurobiological phenomena, explaining mechanisms ranging from discrete cognitive decisions to continuous spatial navigation.

Attractors characterize the long-term states toward which system trajectories evolve, ranging from simple fixed points to complex geometric structures in high-dimensional phase spaces \cite{bib14, bib18, bib19, bib20}. The stability and basin of attraction of these states are key determinants of a neural network’s capacity for robust information processing and memory retention. In attractor-based neural models, information is stored as stable fixed points, allowing networks to reliably converge to predefined states based on initial conditions, a mechanism closely mirroring biological memory formation and retrieval. Understanding attractor dynamics thus offers critical insights into the computational principles underlying neural information processing and memory consolidation.

Attractors have also shown great potential in the optimization and performance enhancement of deep learning models \cite{bib14, bib18, bib32}. By conducting in-depth research on attractors in deep learning models, we can better understand the training process and generalization ability of the models, which in turn provides strong guidance for model improvement. In recurrent neural networks, attractor dynamics are closely related to the expressive power of the network. Different types of attractors correspond to different computational capabilities and information processing methods, which determine the network's performance in tasks such as time-series data processing.

Recent advancements in attractor dynamics have led to substantial theoretical and empirical progress across multiple domains. In neural network research, extensive efforts have been dedicated to elucidating attractor properties across various architectures. For linear threshold neural networks, parameterized models have been employed to systematically characterize continuous attractors, leading to the derivation of precise mathematical criteria for the coexistence of non-degenerate and degenerate equilibrium states \cite{bib5, bib6}. These findings provide fundamental insights into network stability and dynamic behavior, offering a theoretical framework for optimizing neural network architectures.

In the context of recurrent neural networks (RNNs), research has established key conditions and analytical formulations governing the emergence of continuous attractors \cite{bib4, bib14, bib15, bib17}. Notably, the structural properties of synaptic connectivity—such as Gaussian-shaped coupling profiles—along with appropriate parameter tuning, have been shown to play a pivotal role in sustaining continuous attractor states. These insights deepen our understanding of the intrinsic computational mechanisms underlying sequential information processing in RNNs.Furthermore, studies on large-scale neural populations have demonstrated that continuous attractors serve as a crucial substrate for neural information processing \cite{bib14, bib15, bib16, bib17}. By enabling biologically plausible simulations of collective neuronal dynamics, these findings provide powerful tools for modeling large-scale brain activity and advancing our comprehension of cognitive and neural computations.

Despite these significant advancements, a fundamental challenge remains: the lack of a rigorous, unified mathematical framework capable of systematically analyzing attractor properties across diverse dynamical systems. Existing methodologies are often tailored to specific system types, limiting their cross-disciplinary applicability and requiring frequent methodological adaptation when transitioning between different attractor models. This fragmentation not only complicates comparative analyses but also hinders theoretical progress in understanding attractor dynamics at a fundamental level. For example, techniques optimized for studying attractors in neural networks are rarely transferable to complex systems research, and vice versa. Addressing this limitation is crucial for advancing the field toward a more cohesive theoretical foundation.

To bridge this gap, we propose a unified mathematical framework leveraging eigen decomposition and singular value decomposition (SVD) of Jacobian matrices. The Jacobian matrix provides a precise local linear approximation of system dynamics, and its spectral decomposition allows for in-depth characterization of key attractor properties, including stability, basin geometry, and transition dynamics. By establishing a generalized analytical approach rooted in linear algebra, this framework offers broad applicability across a wide range of dynamical systems, laying the groundwork for a more systematic and integrative understanding of attractor behavior.

Our approach offers a unified framework for interpreting previously disparate findings in attractor research. In neural networks, a detailed spectral decomposition of Jacobian matrices provides precise characterizations of equilibrium stability and attractor formation mechanisms\cite{bib5, bib6, bib7}. The eigenvalue magnitudes and signs determine the stability of equilibrium states, while eigenvector orientations delineate the local dynamical evolution, collectively governing the emergence of attractors. Beyond neural systems, this method enables a deeper understanding of phase coexistence and state transitions in complex systems. By analyzing parametric variations in the Jacobian structure, we can systematically identify critical thresholds that drive transitions between coexisting dynamical states\cite{bib10, bib11, bib21, bib22, bib23}. Moreover, in piecewise-smooth systems, our framework provides a powerful tool for dissecting bifurcation pathways from stable fixed points to multi-attractor regimes, offering direct insights into the mechanisms underlying these transitions\cite{bib6, bib9, bib12, bib13, bib24}.

Additionally, our approach contributes to the validation of the manifold hypothesis, which posits that high-dimensional data often reside on low-dimensional manifolds with structured geometric properties. The intrinsic link between attractor dynamics and data manifolds becomes evident through our analysis, revealing how attractors encode low-dimensional representations within high-dimensional state spaces\cite{bib1, bib4, bib9, bib25, bib26}. By examining attractor distributions and their evolution along these manifolds, our findings provide both theoretical reinforcement and empirical support for this foundational hypothesis in neural computation and high-dimensional data analysis.

Experimental validation across multiple dynamical systems substantiates the effectiveness of our proposed framework. Through carefully designed case studies spanning diverse domains, our method demonstrates precise agreement with prior findings while extending analytical capabilities beyond conventional approaches. The results reinforce the robustness and general applicability of this framework, providing a solid empirical foundation for attractor analysis.

Future research directions include applying this methodology to deep neural networks (DNNs), where approximate continuous attractors frequently emerge during training. A systematic examination of Jacobian structures could unveil critical insights into optimization landscapes, generalization mechanisms, and training stability, addressing fundamental challenges in DNN development. Beyond performance enhancement, a deeper understanding of hierarchical attractor dynamics in artificial intelligence systems may yield groundbreaking perspectives on the origins of intelligence. This conceptual shift—from merely simulating intelligent behavior to deciphering its underlying principles—could redefine AI research paradigms.

Given the fundamental role of attractors in both biological cognition and artificial intelligence, our proposed mathematical framework offers a unified perspective that bridges these domains. By establishing a rigorous and generalizable approach, this work opens new avenues for transformative advancements in neuroscience, machine learning, and beyond, potentially driving the next generation of research in intelligent information processing.

\section{Preliminaries}\label{sec2}
\subsection{Attractors of Recursive Neural Networks}
Consider the following recursive neural network model:

\begin{equation}
\dot{x}=\sigma(Wx(t)+b)-Ax(t),\label{eq1}
\end{equation}
where $t\geq 0,x=(x_1,\cdots,x_n)^T\in R^n$ represents the state of the neural network,$W\in R^n$is the weight matrix,$b=(b_1,\cdots,b_n)$is an external input, and $\sigma$ is a continuous function that is almost everywhere differentiable and satisfies the Lipschitz condition.

\begin{definition}[Equilibrium Point\cite{bib27}\cite{bib28}]\label{def1}
A vector $x^\ast\in R^n$ is called an equilibrium point of $\left(\ref{eq1}\right)$ if it satisfies 
\begin{equation}\label{eq:equilibrium}
    \sigma(Wx^* + b) - Ax^* = 0
\end{equation}
\end{definition}

\begin{definition}[Stable Equilibrium Point\cite{bib27}\cite{bib28}]\label{def2}
An equilibrium point $x^\ast$ is stable if for every $\epsilon>0$, there exists a $\delta>0$ such that for every initial condition $x\left(0\right)$ satisfying
$$\left|\left|x\left(0\right)-x^\ast\right|\right|\le\delta,$$
it holds that
$$|\left|x\left(t\right)-x^\ast\right||\le\epsilon \quad \text{ for all t} \geq0.$$
\end{definition}

\begin{definition}[Continuous Attractor\cite{bib27}\cite{bib28}]\label{def3}
A set of equilibrium points $C$ is called a continuous attractor if every $x^\ast\in C$ is stable.
\end{definition}

\subsection{Background on Differential Manifolds}\label{subsec22}

To provide a rigorous mathematical foundation for the framework, it is necessary to introduce some relevant mathematical knowledge form manifold theory. However, to prevent the main thread from getting lost in mathematical definitions, we will fully explain the purpose of introducing each definition and how it contributes to our framework.

To introduce the concept of manifolds, it is necessary to explain topological spaces:

\begin{definition}[Topological Space\cite{bib29}]\label{def4}
Let $X$ be a non-empty set, and let $\mathfrak{I}$ be a collection of subsets of $X$ satisfying the following conditions:

1. $X,\ \emptyset\in\mathfrak{I}$,

2. If $A,\ B\in\mathfrak{I}$, then $A\cap B\in\mathfrak{I}$,

3. If $\mathfrak{I}_1\subset\mathfrak{I}$, then $\bigcup_{A\in\mathfrak{I}_1} A\in\mathfrak{I}$,

then $\mathfrak{I}$ is called a topology on X, and the pair$(X,\mathfrak{I})$ is referred to as a topological space. Each element of $\mathfrak{I}$ is called an open set.

\textup{In brief, a topological space fundamentally defines what open sets and closed sets are, with open sets being the core elements that satisfy the three topological axioms.}
\end{definition}

In neural networks, their expressions are often continuous, which brings convenience to research. Therefore, it is necessary to introduce continuous mappings in manifolds:
\begin{definition}[Continuous Mapping\cite{bib30}]\label{def5}
Let $X$ and $Y$ be two topological spaces, and let $f:X\rightarrow Y$, be a function. If for every open set $U\subseteq Y$, the preimage $f^{-1}(U)$ is an open set in $X$, then $f$ is called a continuous map from $X$ to $Y$.
\end{definition}

To study the dimension of attractors, we need a method to measure their dimension. Therefore, we introduce the concept of homeomorphism, which measures the local Euclidean dimension:

\begin{definition}[Homeomorphism\cite{bib29}]\label{def6}
Let $X$ and $Y$ be two topological spaces. If $f:X\rightarrow Y$ is a one-to-one mapping such that both $f$ and its inverse $f^{-1}$ are continuous, then $f$ is called a homeomorphism from $X$ to $Y$.
\end{definition}

With the above concepts, we can define a manifold rigorously:

\begin{definition}[Topological Manifold\cite{bib29}]\label{def7}
Let $M$ be a Hausdorff space with a countable basis. For every point $p\in M$, there exists an open neighborhood $U_p$ of $p$ and a homeomorphism $\varphi_p:U_p\rightarrow V$, where $V$ is an open subset $R^n$. Then $M$ is called an $n$-dimensional topological manifold. The open neighborhood $U_p$ is referred to as a local coordinate neighborhood, and the pair $(U_p,\varphi_p)$ is called a local coordinate chart or coordinate chart.

\textup{It should be specially noted that in manifold theory, the coordinates of a point and the point itself are two distinct concepts. They are related through what is called a chart(a homeomorphism), which can be understood as a local coordinate system. A chart assigns coordinates to the points within it.}
\end{definition}

To perform calculus-related computations on a manifold, it is necessary to introduce a differential structure. When encountering this concept for the first time, one may have difficulty understanding it. It is sufficient to know that it represents k-th order differentiability:

\begin{definition}[$C^k$-Differentiable Manifold\cite{bib30}]\label{def8}
A $C^k$-differentiable structure on a topological manifold $M$ (where $1\le k\le\infty$) is a collection of local coordinate charts
$$\Phi={\left(U_\alpha,\varphi_\alpha\right):\alpha\in A},$$
(where $A$ is an indexing set) that satisfies the following conditions:

1. $M=\bigcup_{\alpha\in A} U_\alpha$,

2. For all $\alpha$,$\beta\in A$, the coordinate charts$\ \left(U_\alpha,\varphi_\alpha\right)$ and $\left(U_\beta,\varphi_\beta\right)$ are $C^k$-compatible.

Specifically, when $W_{\alpha\beta}=U_\alpha\cap U_\beta\neq\emptyset$, the transition maps $\varphi_\beta\circ\varphi_\alpha^{-1}$ and $\varphi_\alpha\circ\varphi_\beta^{-1}$ are $C^k$-class diffeomorphisms between the open subsets $\varphi_\alpha\left(W_{\alpha\beta}\right)$ and $\varphi_\beta\left(W_{\alpha\beta}\right)$ in $R^n$.
The collection $\Phi$ is maximal with respect to the compatibility condition. This means that if there exists a local coordinate chart $(U,\varphi)$ that is $C^k$-compatible with every chart in $\Phi$, then $(U,\varphi)$ is already included in $\Phi$. 

An $n$-dimensional manifold $M$, together with its $C^k$- differentiable structure $\Phi$, denoted as $(M,\Phi)$, is called an $n$-dimensional $C^k$-differentiable manifold, abbreviated as a $C^k$-differentiable manifold. When the transition maps $\varphi_\beta\circ\varphi_\alpha^{-1}$ and $\varphi_\alpha\circ\varphi_\beta^{-1}$ are $C^\infty$-class diffeomorphisms, $M$ is called a smooth manifold, and each coordinate chart in $\Phi$ is referred to as a smooth coordinate chart.
\end{definition}

A smooth mapping corresponds to infinite-order differentiability:

\begin{definition}[Smooth Map\cite{bib30}]\label{def9}
Let $M$, $N$ be smooth manifolds, and let $f:M\rightarrow N$ be a mapping. For a point $p\in M$, if there exists a smooth coordinate chart $\left(U,\varphi\right)$ on $M$ and a smooth coordinate chart $(V,\psi)$ on $N$ such that:
1. $p\in U, f(U)\subseteq V$,
2. the composition $\psi\circ f\circ\varphi^{-1}:\varphi(U)\rightarrow\psi(V)$is smooth at $\varphi(p)$,
then $f$ is said to be smooth at the point $p$. If f is smooth at every point in $M$, then $f$ is called a smooth map from $M$ to $N$.
\end{definition}

The core of our framework is based on linear approximation, so it is necessary to introduce tangent vectors and tangent spaces:

\begin{definition}[Tangent Space\cite{bib30}]\label{def10}
Let \( M \) be an \( n \)-dimensional smooth manifold. For \( p \in M \), denote \( C^{\infty}(p) \) as the set of smooth functions defined in some neighborhood of \( p \). A function \( X_p: C^{\infty}(p) \to \mathbb{R} \) is called a tangent vector to \( M \) at \( p \) if and only if for all \( f, g \in C^{\infty}(p) \), the following conditions are satisfied:

1. If there exists a neighborhood \( U \) of \( p \) in \( M \) such that \( f|_U = g|_U \) (where \( f|_U \) denotes the restriction of \( f \) to \( U \)), then \( X_p(f) = X_p(g) \).

2. \(\forall  \alpha, \beta \in \mathbb{R} \),
\[
X_p(\alpha f + \beta g) = \alpha X_p(f) + \beta X_p(g),
\]
    where the operation on functions is defined as
\[
(\alpha f + \beta g)(q) = \alpha f(q) + \beta g(q).
\]

3.
\[
X_p(f \cdot g) = f(p) X_p(g) + g(p) X_p(f),
\]
    where
\[
(f \cdot g)(q) = f(q) g(q).
\]

The set of all tangent vectors to \( M \) at \( p \) is denoted by \( T_p M \). For \( X_p, Y_p \in T_p M \), \( \alpha \in \mathbb{R} \), and \( f \in C^{\infty}(p) \), we define
\[
(X_p + Y_p)(f) = X_p(f) + Y_p(f),
\]
\[
(\alpha X_p)(f) = \alpha X_p(f).
\]

Thus, \( T_p M \) is a vector space, called the tangent space to \( M \) at \( p \). It can be shown that the dimension of the tangent space is equal to the dimension of the manifold.
\end{definition}

We use the rank of the local linear approximation of a mapping to definite its rank:

\begin{definition}[Rank of a Linear Map\cite{bib30}]\label{def11}
Consider a linear map $F:X\rightarrow Y$, where $X$ and $Y$ are vector spaces of dimensions $m$ and $n$, respectively. Let ${\delta_1,\cdots,\delta_m}$ and ${\varepsilon_1,\cdots,\varepsilon_n}$be bases for $X$ and $Y$, respectively. The image of each basis vector $F(\delta_i)$ in $Y$ can be expressed in terms of the basis ${\varepsilon_1,\cdots,\varepsilon_n}$ as:
$$F\left(\delta_i\right)=\left(\varepsilon_1,\cdots,\varepsilon_n\right)\left(\begin{array}{c}
    a_i^1  \\
     \vdots\\
     a_i^n
\end{array}\right).$$
Thus, the linear map F can be represented as:
\begin{align*}
F\left(\delta_1,\cdots,\delta_m\right) &= \left(F\left(\delta_1\right),\cdots,F\left(\delta_m\right)\right) \\
&= \left(\varepsilon_1,\cdots,\varepsilon_n\right)
\left(\begin{matrix}
a_1^1 & \cdots & a_m^1 \\
\vdots & \ddots & \vdots \\
a_1^n & \cdots & a_m^n \\
\end{matrix}
\right) \\
&= \left(\varepsilon_1,\cdots,\varepsilon_n\right)A.
\end{align*}
Define the rank of the linear map $F$ as the rank of the matrix $A$, denoted as $rank\ F$. From linear algebra, it follows that:
$$rank\ F=\dim F(X).$$
\end{definition}

\begin{definition}[Rank of a Smooth Map\cite{bib30}]\label{def12}
Let $M$ and $N$ be smooth manifolds of dimensions $m$ and $n$, respectively, and let $F:\ M\rightarrow N$ be a smooth map. For a point $p\in M$, the differential of $F$ at $p$, denoted by $F_{\ast p}$, is a linear map from the m-dimensional tangent space $T_pM$ to the $n$-dimensional tangent space $T_{F(p)}N$:

$$F_{\ast p}:\ T_pM\rightarrow T_{F(p)}N.$$

The rank of the differential $F_{\ast p}$ is referred to as the rank of the smooth map $F$ at the point $p$, denoted by $rank_p F$.

Let $\left(U,\varphi,x^i\right)$ and $\left(V,\psi,y^j\right)$ be local coordinate charts for $M$ and $N$ at points $p$ and $q=F(p)$, respectively. Assume that $F(U)\subset V$. It can be readily shown that the matrix representation of the differential $F_{\ast p}$ with respect to the bases $\left\{\frac{\partial}{\partial x^i}\right\}\subset T_p M$ and $\left\{\frac{\partial}{\partial y^j}\right\}\subset T_p N$ is equal to the Jacobian matrix of the local representation $\hat{F}=\psi\circ F\circ\varphi^{-1}$. Therefore:

$$rank_p F=rank\left(\left(\frac{\partial{\hat{F}}^j}{\partial x^i}\right)_{\varphi\left(p\right)}\right).$$

It is evident that:

$$rank_p F\le\min(m,n).$$
\end{definition}

For a subset of a manifold, whether it can form a lower-dimensional manifold will be addressed using a lemma in the following text. To this end, we need to introduce the concept of the property of k-dimensional Submanifold:

\begin{definition}[Property of $k$-Dimensional Submanifold\cite{bib30}]\label{def13}
Let $N$ be an $n$-dimensional smooth manifold, and let $A$ be a subset of $N$. The subset $A$ is said to possess the property of being a $k$-dimensional submanifold of $N$ if and only if for every point $q\in A$, there exists a coordinate chart $\left(V,\psi\right)$ of $N$ at q satisfying the following conditions:
$$\psi\left(q\right)=0, \psi\left(V\right)=C_\varepsilon^n\left(0\right), \psi\left(V\cap A\right)=C_\varepsilon^k\left(0\right)\times\left\{0\right\}^{n-k},$$
where $C_\varepsilon^n\left(0\right)$ denotes the $\varepsilon$-neighborhood of the origin in $\mathbb{R}^n$.
\end{definition}

\section{Attractor Dimension Theorem}\label{sec3}

To facilitate the proof of subsequent theorems, we introduce two lemmas here. The first lemma establishes a relation between the rank of a smooth map and local coordinate systems. The second lemma provides a sufficient condition for determining the dimension of submanifolds, a condition that is strongly connected with the dimension of attractors. 

\begin{lemma}[Rank Theorem\cite{bib30}]\label{lem1}
Let $M$ and $N$ be smooth manifolds of dimensions $m$ and $n$, respectively. Let $F:M\rightarrow N$ be a smooth map. Suppose that the point $p\in M$ has an open neighborhood $W$ in which the rank of $F$ is constantly $k$. Then, there exist local coordinate charts $\left(U,\varphi,x^i\right)$ for $M$ at $p$ and $\left(V,\psi,y^j\right)$ for $N$ at $q=F(p)$, with $U\subset W$, such that the local representation of $F$ is:

$$\hat{F}=\psi\circ F\circ\varphi:(x^1,\cdots,x^m)\mapsto(x^1,\cdots,x^k,0,\cdots,0).$$

\end{lemma}

\begingroup
\let\originalcite\cite
\renewcommand{\cite}[1]{\unskip\originalcite{#1}}
\begin{lemma}[\cite{bib30}]\label{lem2}
Let $N$ be a smooth manifold of dimensions $n$, and let A be a subset of $N$ that has the property of $k$-Dimensional Submanifold. Then A is a $k$-Dimensional differentiable manifold, where the topology of $A$ is the subspace topology inherited from $N$, and the differentiable structure is generated by a specific atlas.
\end{lemma}
\endgroup

\vspace{12pt} 

\begin{theorem}[Submanifold Dimension Theorem]\label{thm1}
Let $M$ and $N$ be smooth manifolds of dimensions $m$ and $n$, respectively. Let $F:M\rightarrow N$ be a smooth map. There exists an open neighborhood $\widetilde{U}$ of $p$ such that $A\cap\widetilde{U}$ is a regular sub-manifold $(m-k)$ dimensional of $M$, if:

1. The rank of $F$ on $M$ is less than or equal to a constant $k$,

2. For some point $q\in N$, there exists a point $p\in A=F^{-1}(q)$ such that the rank of $F$ at $p$ is exactly $k$.

\end{theorem}

\vspace{12pt} 

\begin{proof}
Choose local coordinate charts $\left(U_1,\varphi_1\right)$ for $M$ at $p$ and $(V_1,\psi_1)$ for $N$ at $q$. In these coordinates, the local representation of $F$ is $\hat{F}=\psi_1\circ F\circ\varphi_1^{-1}$. Since the Jacobian matrix of $\hat{F}$ has a $k\times k$ minor $J_k(\rho)$ that does not vanish at $\rho=r$, by continuity, there exists an open neighborhood $\widetilde{U}\subset U_1$ around $p$ where${\ J}_k(\rho)\neq0$. Therefore, the rank of $F$ on $\widetilde{U}$ is at least $k$. By \textit{assumption 1}, the rank of $F$ on $\widetilde{U}$ is exactly $k$.

$\widetilde{U}$ is an open subset of $M$, and hence an open $m-$dimensional submanifold of $M$. The restriction $f=F|_{\widetilde{U}}:\widetilde{U}\rightarrow N$ is a smooth map between smooth manifolds with constant rank $k$.

To show that $A\cap\widetilde{U}$ is an $(m-k)$-dimensional regular submanifold of $M$:
Consider any $s\in A\cap\widetilde{U}$, then $f\left(s\right)=q$. By Lemma \ref{lem1}, there exist local coordinate charts $\left(U_2,\varphi_2,x_i\right)$ for $\widetilde{U}$ at $s$ and $(V_2,\psi_2,y_j)$ for $N$ at $q$ such that:

$$\varphi_2\left(s\right)=0, \psi_2\left(q\right)=0,$$
$$\varphi_2\left(U\right)=C_\varepsilon^m\left(0\right),\psi_2\left(V\right)=C_\varepsilon^n\left(0\right),$$
$$\hat{f}=\psi_2\circ f\circ\varphi_2^{-1}:C_\varepsilon^m\left(0\right)\rightarrow C_\varepsilon^n\left(0\right),$$
$$(x_1,\cdots,x_m)\rightarrow(x_1,\cdots,x_k,0,\cdots,0).$$

For any $t\in U_2\cap A$:
$$\varphi_2\left(t\right)=\left(x_1,\cdots,x_m\right)\in\varphi_2\left(U\right)=C_\varepsilon^m\left(0\right),$$
\begin{align*}
f(t) = q \Longrightarrow \psi_2(q) &= 0 \\
&= \psi_2 \circ f \circ \varphi_2^{-1}(x_1,\cdots,x_m) \\
&= (x_1,\cdots,x_k,0,\cdots,0).
\end{align*}

Therefore:
$$x_1=\cdots=x_k=0.$$

Hence, $t\in U_2\cap A$ if and only if:
$$\varphi_2\left(t\right)=(0,\cdots0,x_{k+1},\cdots,x_m).$$

Therefore:
$$\varphi_2\left(U_2\cap A\right)=\left\{0\right\}^k\times C_\varepsilon^{m-k}\left(0\right).$$

Thus, for every $s\in A\cap\widetilde{U}$, there exists a coordinate chart $\left(U_2,\varphi_2,x_i\right)$ for $M$ at $s$ satisfying:
$$\varphi_2\left(s\right)=0,$$
$$ \varphi_2\left(U\right)=C_\varepsilon^m\left(0\right),$$ 
$$\varphi_2\left(U_2\cap A\cap\widetilde{U}\right)=\varphi_2\left(U_2\cap A\right)=C_\varepsilon^{m-k}\left(0\right)\times\left\{0\right\}^k.
$$

Therefore, $A\cap\widetilde{U}$ possesses the structure of an $(m-k)$-dimensional manifold. By Lemma \ref{lem2}, $A\cap\widetilde{U}$ is an $(m-k)$-dimensional submanifold of $\widetilde{U}$.
\end{proof}

\vspace{12pt} 

Using theorem \ref{thm1} in network \ref{eq1}, we obtain the theorem below: 

\begin{theorem}[Attractor Dimension Theorem]\label{thm2}
Consider the network defined by:
$$\dot{x}=\sigma(Wx+b)-Ax,$$
where $\sigma\left(Wx+b\right)-Ax$ is a smooth function. Suppose that:

1. $rank\left(\frac{\partial\sigma\left(Wx+b\right)-Ax}{\partial x}\right)\le r$, $\forall x\in\mathbb{R}^n$,

2. There exists $x_0\in X$ s.t. $rank\left(\frac{\partial\sigma\left(Wx+b\right)-Ax}{\partial x}\right)|_{x_0}=r$, where $X$ is the set of solutions to the equation.

Then there exists an $(n-r)$-dimensional attractor manifold in the vicinity of $x_0$.
\end{theorem}

\vspace{12pt} 

\begin{proof}
The equilibrium points of the network satisfy:
\begin{equation}\label{eq:equilibrium}
\sigma(Wx+b)-Ax=0.
\end{equation}

These are the zeros of the map $F:\mathbb{R}^n\rightarrow\mathbb{R}^n$, defined by
\begin{equation}\label{eq:Fx_definition}
F(x) = \sigma(Wx + b) - Ax
\end{equation} 

Letting $q=0$ in Theorem \ref{thm1}, we have \begin{equation}\label{eq:Fx_definition} 
F^{-1}\left(q\right)=X.
\end{equation} 

By Theorem \ref{thm1}, the theorem holds.
\end{proof}

It should be noted that we only focus on the dimension of the manifold, which means the aforementioned attractor manifold may not be a true attractor. However, this is not particularly important since there are many existing techniques to analyze the stability of manifold such as the Lyapunov methods.

\vspace{12pt} 

\begin{theorem}[Attractor Dimension Estimate]\label{thm3}
Consider the system:
$$\dot{x}=\sigma\left(Wx+b\right)-Ax=\begin{pmatrix}\sigma_1(W_1x+b_1)-A_1x\\\vdots\\\sigma_n(W_nx+b_n)-A_nx\end{pmatrix},$$
where $W_i$ and $A_i$ denote the $i$-th rows of the matrices $W$ and $A$, respectively.

Let $f_i(x)=\sigma_i\left(W_ix+b_i\right)-A_ix$. The function space $\{f_1,\cdots,f_n\}$ forms a vector space. If $\{f_1,\cdots,f_n\}$ is linearly dependent in $\mathbb{R}^n$, and the number of vectors in the maximal linearly independent subset is $k$, and there exists $x_0\in X$ such that $rank\left(\frac{\partial\sigma\left(Wx+b\right)-Ax}{\partial x}\right)|_{x_0}=k$, then the attractor dimension satisfies $\dim X=n-k$.
\end{theorem}

\vspace{12pt} 

\begin{proof}
If $\{f_1,\cdots,f_n\}$ is linearly dependent in $\mathbb{R}^n$, then ${\frac{\partial f_1}{\partial x},\cdots,\frac{\partial f_n}{\partial x}}$ also linearly dependent in $\mathbb{R}^n$.
Moreover, the number of vectors in the maximal linearly independent subset is $k$. Therefore,
$$rank\left(\frac{\partial\sigma\left(Wx+b\right)-Ax}{\partial x}\right)\le k.$$

By the premises of Theorem \ref{thm2}, there exists an $(n-r)$-dimensional attractor manifold.

This theorem is particularly useful when $\sigma$ is a linear function or a piecewise linear function (e.g., $ReLU$).

\section{Theorem Verification}\label{sec4}

To verify the compatibility between our conclusions and the relevant conclusions from previous studies, we will use the conclusions derived from this paper to conduct a rigorous validation of the results obtained in the past\cite{bib28}.

From existing literature, the following conclusion is known:

Consider the recurrent neural network model:
\begin{equation}\label{eq:relu_rnn}
\dot{x} = -x + Wf(x(t)) + b
\end{equation}
where $f$ is $ReLU$ function.

Let $P,Z\subseteq\{1,2,\cdots,n\}$ be index sets containing $p$ and $z$ elements, respectively, with $P\cap Z=\emptyset$ and $P\cup Z=\{1,2,\cdots,n\}$. Thus, the network can be represented as:
$$\begin{bmatrix}\dot{x}_P(t)\\\dot{x}_Z(t)\end{bmatrix}+\begin{bmatrix}x_P(t)\\x_Z(t)\end{bmatrix}=\begin{bmatrix}W_P&W_{PZ}\\W_{ZP}&W_Z\end{bmatrix}\cdot f\begin{bmatrix}x_P(t)\\x_Z(t)\end{bmatrix}+\begin{bmatrix}b_P\\b_Z\end{bmatrix},$$
where $W_P$ and $W_Z$ denote the respective submatrices of $W$, and similarly for $A$.
Let $\lambda_i^P(i=1,\cdots,p)$ be the eigenvalues of $W_P$, and $v_i^P\left(i=1,\cdots,p\right)$ the corresponding unit eigenvectors. If $W_P$ is symmetric, the following theorem holds:

Assume that $\lambda_i^P=1$ is the largest eigenvalue of $W_P$ with geometric multiplicity m, and $b_P=0$. Then the set

$$C=\left\{\begin{bmatrix}x_P^*=\sum_{i=1}^mc_iv_i^P\\x_Z^*=W_{ZP}\cdot x_p^*+b_Z\end{bmatrix}\right\}$$
is a continuous attractor, where $x^\ast=\left(x_P^\ast,x_Z^\ast\right)^T$ satisfies:
$$\begin{cases}(x_P^*)_k\geq0\\(x_Z^*)_k<0\end{cases}$$
This network model differs slightly from the previously discussed model. However, similar to how Theorem \ref{thm2} was derived from Theorem \ref{thm1}, we only need to verify $rank(\frac{\partial(-x+Wf\left(x\left(t\right)\right)+b)}{\partial x})$. Before doing so, it is necessary to address the $ReLU$ function. The $ReLU$ function is not differentiable at 0, but this non-differentiability can be ignored from a measure-theoretic perspective if the set of equilibrium points forms a continuous attractor, as such points constitute a set of measure zero.

Now, we will use the results of this paper to prove this previous conclusion. 

consider $rank\left(\frac{\partial\left(-x+Wf\left(x\left(t\right)\right)+b\right)}{\partial x}\right)$:
\begin{equation}\label{eq:jacobian_relu}
\frac{\partial (-x + Wf(x(t)) + b)}{\partial x} = \frac{\partial Wf(x(t))}{\partial x} - I
\end{equation}

where $f$ is $ReLU$ function. Therefore:

$$\frac{\partial Wf\big(x(t)\big)}{\partial x}=\frac{\partial\big[\begin{matrix}W_Px_P\\W_ZPx_P\end{matrix}\big]}{\partial x}=\begin{bmatrix}W_P&0\\W_{ZP}&0\end{bmatrix}.$$

Since $W_P$ is symmetric and its largest eigenvalue is 1 with geometric multiplicity $m$, it can be diagonalized as:
\begin{equation}\label{eq:eigendecomp} 
W_P = V \Lambda V^{-1}
\end{equation}

where $\Lambda $ is a diagonal matrix with $m$ entries equal to 1 and the remaining entries less than 1. Thus:
$$\frac{\partial Wf\left(x\left(t\right)\right)}{\partial x}-I=\left[\begin{matrix}V\left(\Lambda-I\right)V^{-1}&0\\W_{ZP}&-I\\\end{matrix}\right]=\left[\begin{matrix}V\Lambda^\ast V^{-1}&0\\W_{ZP}&-I\\\end{matrix}\right],$$
where $\Lambda^\ast$ is a diagonal matrix with $m$ entries equal to 0 and the remaining entries less than 0.

Consequently, $V\Lambda^\ast V^{-1}$ has m rows of all zeros, and therefore, the matrix $\left[\begin{matrix}V\Lambda^\ast V^{-1}&0\\W_{ZP}&-I\\\end{matrix}\right]$ also has $m$ rows of all zeros. Hence:
\begin{equation}\label{eq:final_rank}
\mathrm{rank} \left( \frac{\partial (-x + Wf(x(t)) + b)}{\partial x} \right) = n - m
\end{equation}
Since $\frac{\partial Wf\left(x\left(t\right)\right)}{\partial x}-I$ is independent of $x$, it is easy to verify that:
$$\exists x_0\in C=\left\{\begin{bmatrix}x_P^*=\sum_{i=1}^mc_iv_i^P\\x_Z^*=W_{ZP}\cdot x_p^*+b_Z\end{bmatrix}\right\},$$
$$s.t. \quad rank\left(\frac{\partial\left(-x+Wf\left(x(t)\right)+b\right)}{\partial x}\right)|_{x_0}=n-m,$$
and 
$$0=-x_0+Wf\left(x_0\right)+b.$$
By Theorem \ref{thm1}, there exists an $n-\left(n-m\right)=m$-dimensional continuous attractor.
\end{proof}

\section{Experiments}\label{sec5}

\begin{figure*}[t!]
    \centering
    \captionsetup[sub]{
                        font = normalsize,
                        labelfont = bf,
                        }
    \subfloat[][\label{fig1:a}] 
        {\includegraphics[width=0.48\textwidth]{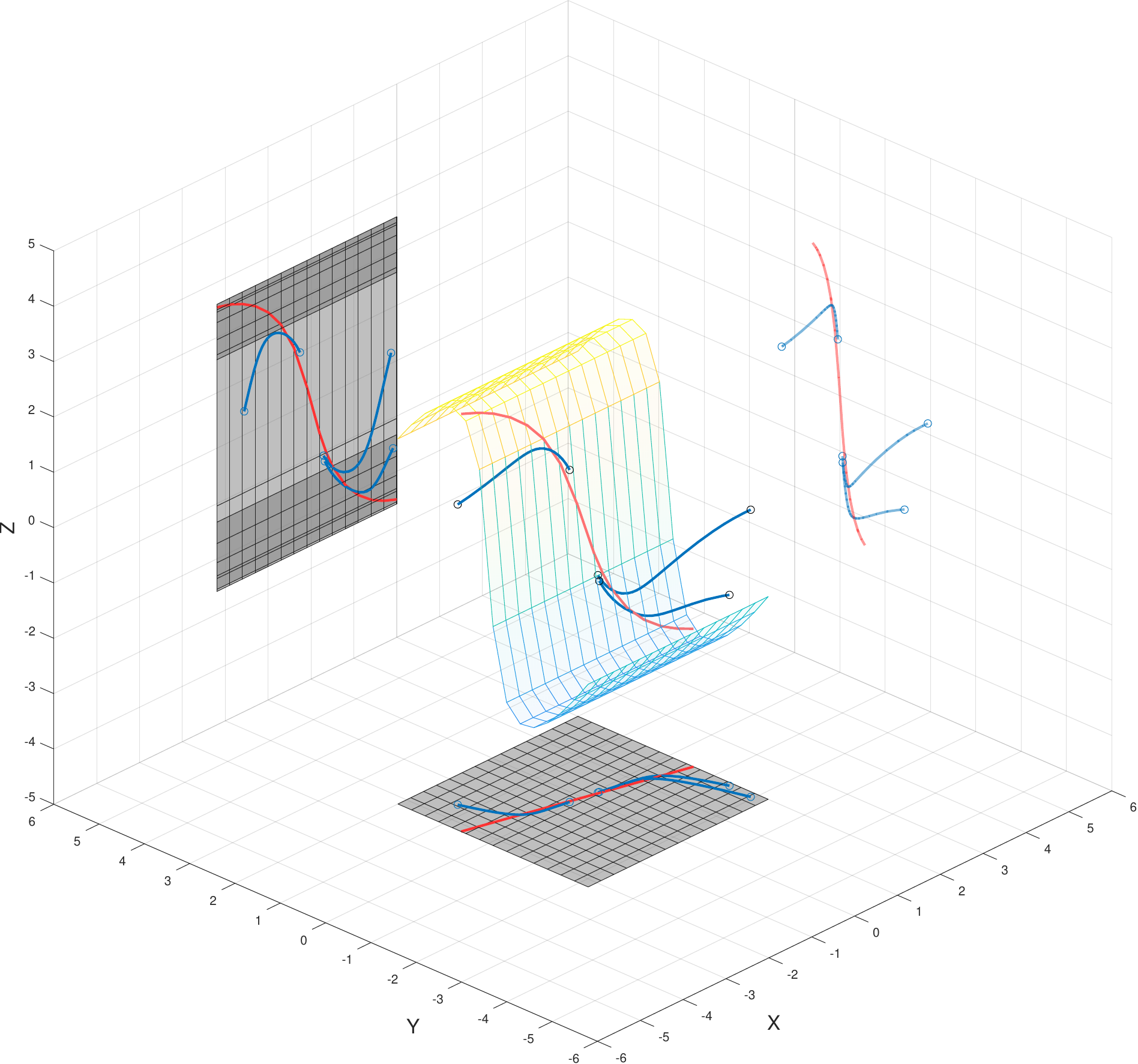}}
    \hfill
    \subfloat[][\label{fig1:b}]
        {\includegraphics[width=0.48\textwidth]{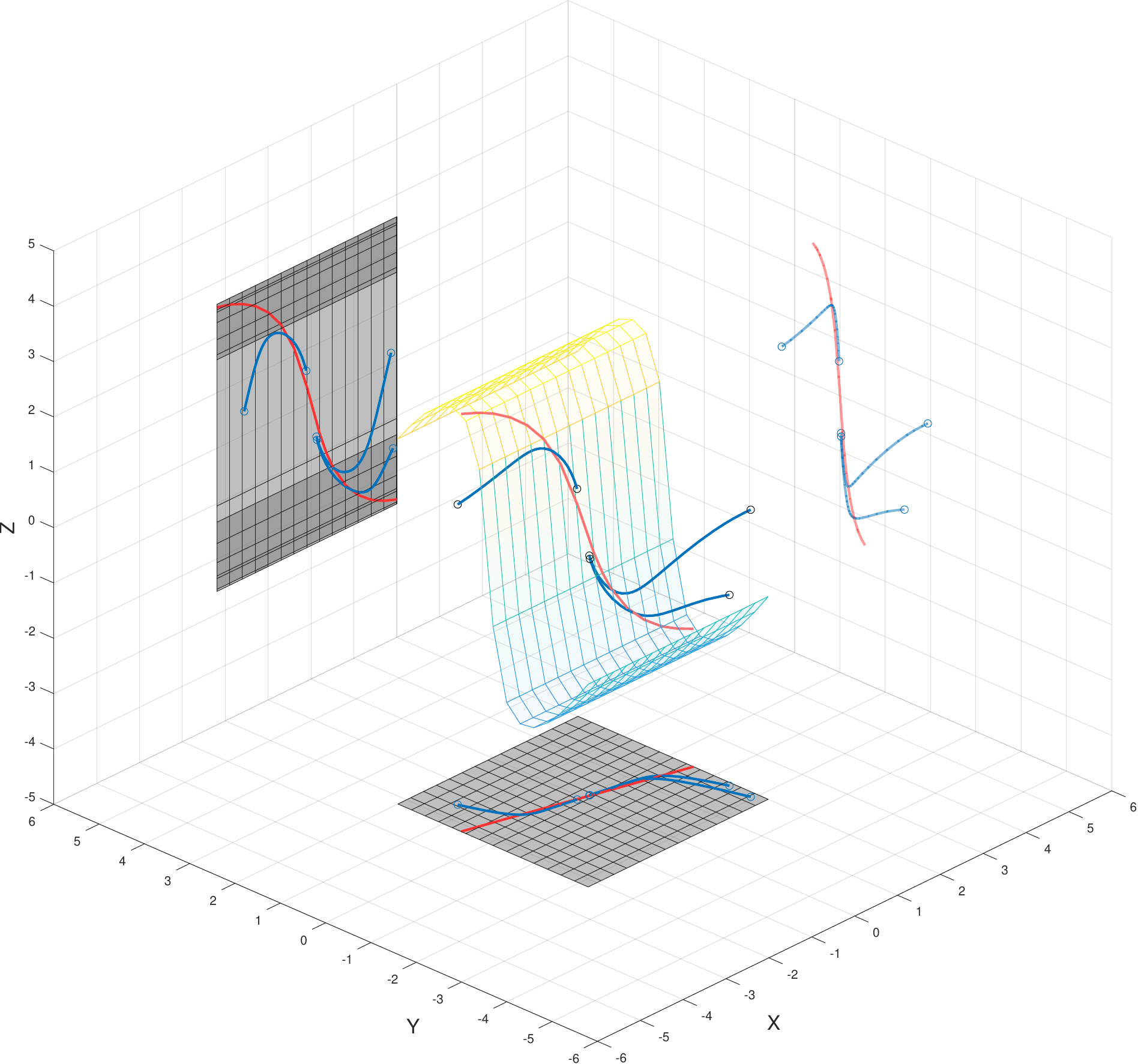}}
    \hfill

    \subfloat[][\label{fig1:c}]
        {\includegraphics[width=0.48\textwidth]{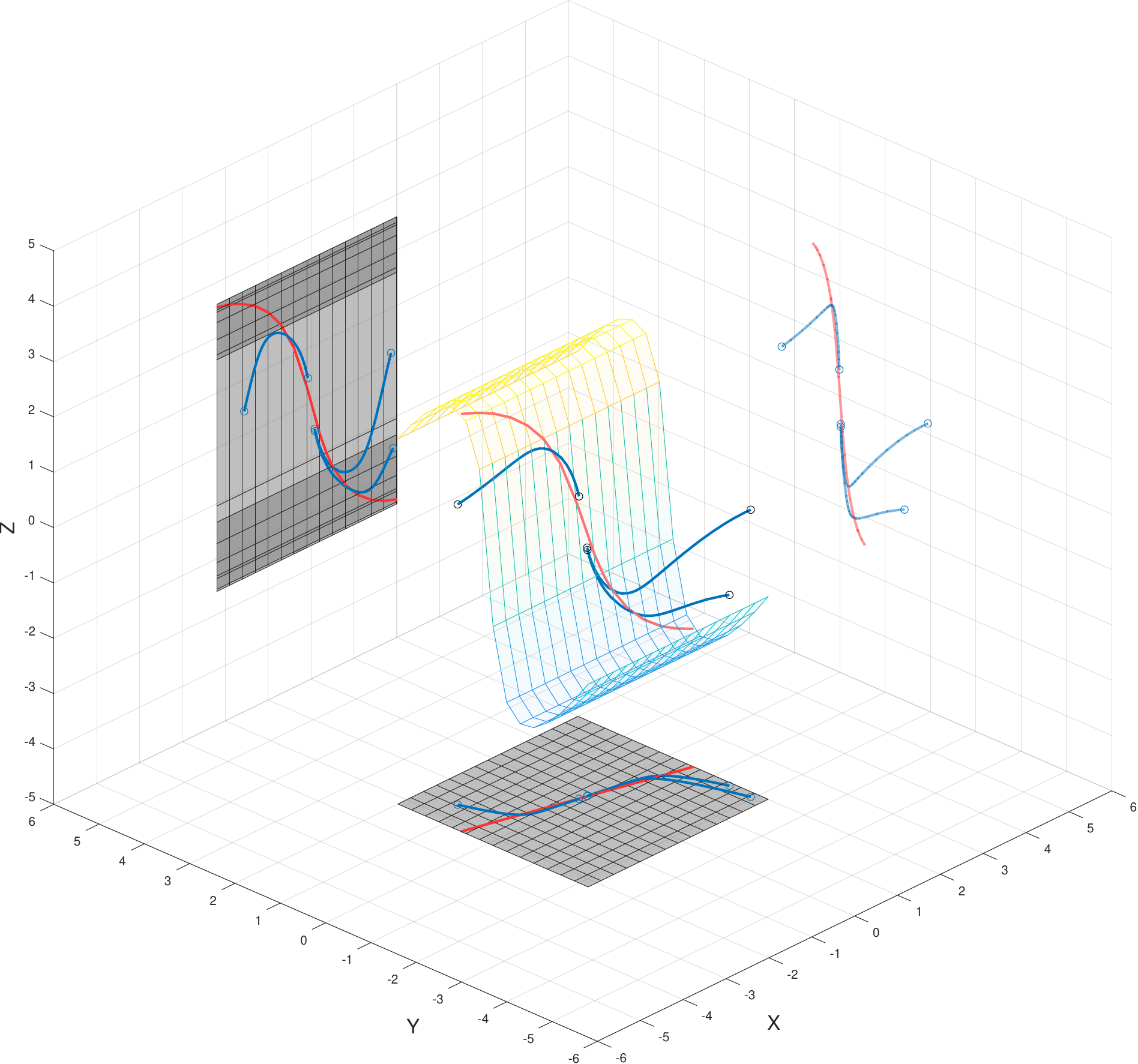}}
    \hfill
    \subfloat[][\label{fig1:d}]
        {\includegraphics[width=0.48\textwidth]{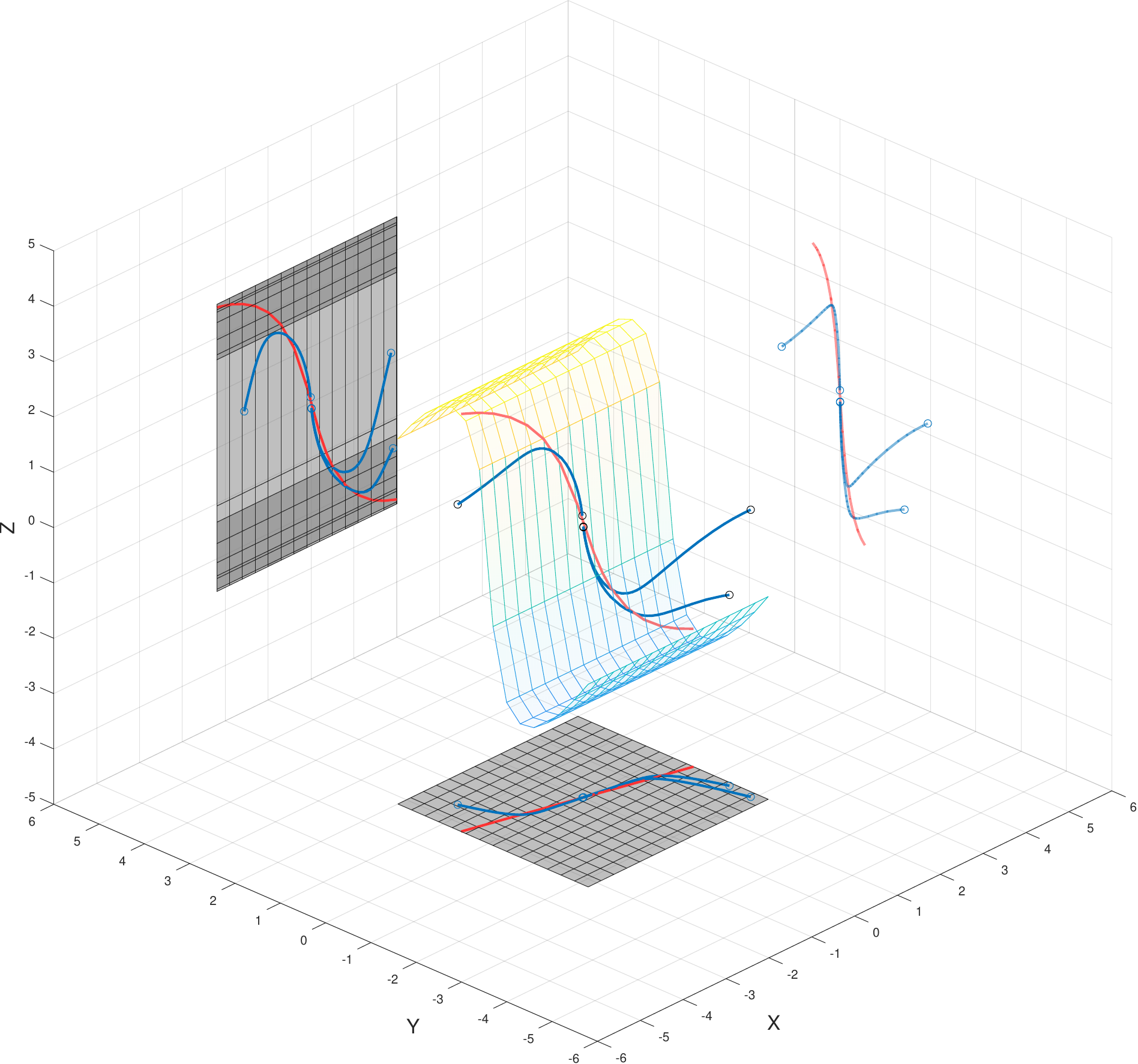}}
    
    \caption{The iteration trajectories of a discrete dynamical system $x(t+1)=\sin(Wx(t)+b)-Ax(t)$ with eigenvalue stratification characteristics at time steps $t=50(a), 100(b), 200(c)$, and $20000(d)$.}
    \label{fig_1}
\end{figure*}

In model \ref{eq1}, continuous attractors are, in fact, solutions to the equation $\sigma(Wx(t) + b) - Ax(t) = 0$ that exhibit Lyapunov stability. In practical applications, due to the randomness of initial parameters and the limitations of current neural network training methods, the equation $\sigma(Wx(t) + b) - Ax(t) = 0$ typically has either a single solution or a finite number of solutions. This leads to the difficulty in observing continuous attractor behavior in current neural networks from a theoretical standpoint. However, examples in theoretical neural network models have demonstrated the occurrence of continuous attractor behaviors \cite{bib5}. In \cite{bib31}, it is suggested that one can perform a slow-fast decomposition of the manifold and linearize the map at specific points, obtaining the Jacobian matrix at those points. The eigenvalues of the Jacobian matrix characterize the rate of change of the system in the directions of the corresponding eigenvectors near that point. If there is a significant difference in the magnitudes of the eigenvalues, which we refer to as the stratification of eigenvalues, approximate continuous attractor behavior emerges. When a zero eigenvalue exists, the system exhibits a true continuous attractor, as derived in the third section. The theory in Section \ref{sec3} is consistent with the approximate continuous attractor behavior highlighted in \cite{bib31}. In cases where the difference in eigenvalue magnitudes is infinite, the smaller eigenvalues can be considered as zero eigenvalues, and the system can be regarded as degenerate with a zero eigenvalue, in which case a true continuous attractor is present. An example of an approximate continuous attractor is illustrated in  Figure \ref{fig_1}. It can be observed that although the system does not have a continuous attractor, its behavior resembles that of a system with a continuous attractor: the system first rapidly converges to a surface and then moves along the red curve on this surface. The evolution of the trajectories in the dynamical system is extremely slow when trajectories approach the point attractor(a continuous attractor can be regarded as an extreme case in this interpretation, where the movement speed on the low-dimensional curve approaches zero). However, it can be seen that the system eventually converges near the point attractor.

\subsection{Theory}

By integrating this paper's theoretical framework with \cite{bib31}, we propose a method for validating the manifold hypothesis. The manifold hypothesis states that meaningful data reside on certain low-dimensional submanifolds within the ambient space \( \mathbb{R}^n \). This is strongly supported by real-world data such as images, speech, and textual information, whose probability distributions are highly concentrated. Inspired by the rigorous theory of continuous attractors in Section \ref{sec3}, and the observation that the presence of an eigenvalue gap induces approximate continuous attractor behavior, we hypothesize that similar phenomena may occur in neural network models applied to real-world tasks. Specifically, we conjecture that there exists a pronounced difference in the magnitudes of eigenvalues, thereby leading to the conclusion that actual data are confined to certain low-dimensional submanifolds. This, in turn, provides an empirical basis for validating the manifold hypothesis and a highly promising analytical approach for neural networks.

To apply the aforementioned theory to practical neural networks, the first challenge is that, except for certain end-to-end models, the input and output dimensions of artificial neural networks are generally mismatched. This implies that the models are not equivalent to the recurrent neural network (RNN) models discussed in Section \ref{sec3} and \cite{bib31}. As a result, the Jacobian matrix of the overall mapping (where the entire model is treated as a mapping) is no longer a square matrix, rendering eigenvalue decomposition infeasible. To address this issue, we propose utilizing singular value decomposition (SVD) instead of eigenvalue decomposition for non-square Jacobian matrices. Mathematically, replacing the decomposition of eigenvalues with the decomposition of singular values and drawing an analogy between the differences in the magnitudes of singular values and eigenvalues is a reasonable approach: Eigenvalue decomposition breaks down the linear transformation represented by a matrix into scaling operations along its eigenvector directions. The magnitude of an eigenvalue indicates the scaling factor applied by the transformation in the direction of its corresponding eigenvector. Similarly, SVD factorizes any linear transformation into a sequence of a rotation/reflection, a scaling operation, and another rotation/reflection. Analogous to the magnitude of eigenvalues, the magnitude of the singular values quantifies the scaling factors applied during the scaling stage to the corresponding right singular vectors. Consequently, we have reason to hypothesize that the observed stratification phenomenon of both singular values and eigenvalue magnitudes suggests the existence of an approximate continuous attractor.

Through the experiment setups described below, we find that the phenomenon of singular value stratification is widespread in the final mappings obtained by neural networks solving classification tasks. This provides compelling empirical evidence in support of the manifold hypothesis.

\subsection{Setup}

\noindent\textbf{Datasets}. In experiment 1, which is shown in Table \ref{tab1}, we use datasets as follow: MINST\cite{bib33}, CIFAR-10\cite{bib34} and Fruits. MNIST is a commonly used dataset which contains a total of 70,000 grayscale images of handwritten digits (0 through 9), each with a resolution of $28 \times 28$ pixels. CIFAR-10 dataset is consists of 60,000 $32 \times 32$ color images distributed across 10 distinct classes: \textit{airplane, automobile, bird, cat, deer, dog, frog, horse, ship,} and \textit{truck}. Fruit dataset is not a open dataset. It is built by scraping online fruit graphs in different forms of expression and resized them to $224 \times 224$. It has 3374 images of 10 classes: \textit{apple, avocado, banana, cherry, kiwi, mango, orange, pineapple, strawberries} and \textit{watermelon}. Experiment 2 only uses Fruit dataset.

\vspace{0.5em}
\noindent\textbf{Implementation details}. For the purpose of verifying our theory, the models used in our experiments only retain the basic architecture necessary for a classification task. For model parameter settings, we adhere to the well-recognized training configurations in machine learning, encompassing weight decay (5e-4), momentum (0.9), and batch size (32). Our optimizer employs SGD. All experiments are implemented on the PyTorch platform with a single NVIDIA RTX GPU.

\vspace{0.5em}
\noindent\textbf{Training and testing details}. When executing the training process of experiment 2, concerning the need for a control experiment, we only use the first 9 classes of the Fruit dataset, leaving the watermelon class unused in training, but involved in the testing process. Two more classes are added in testing as well: noise image and noise. The noise image class is randomly selected from the ImageNet dataset, which has the same resolution as the Fruit dataset we built. The noise class is $224 \times 224$ images randomly generated by code, with each pixel being a random RGB value. Hence, when testing a model, we encounter four distinct scenarios: (1) the first 9 classes from the Fruit dataset, which were seen during training; (2) the unseen watermelon class, which is consistent with the training classes at the data format level and has semantical similarity; (3) a class of natural images (e.g., non-fruit objects); (4) a noise class comprising completely random pixels that are semantically meaningless. The model evaluation was performed at its optimal performance level.

\textbf{}
\begin{table}[!t]
\caption{Coefficient of Variation of Sample Singular Values}
\label{tab1}
\centering
\resizebox{\textwidth}{!}{
\begin{tabular}{@{}l*{5}{D{.}{.}{2.4}}@{}}
\toprule
 & \multicolumn{5}{c}{\textbf{Model + Dataset}} \\
\cmidrule(lr){2-6}
\raisebox{0.7ex}[0pt]{\textbf{Class}} & \multicolumn{1}{c}{\textbf{MLP+MNIST}} & \multicolumn{1}{c}{\textbf{CNN+MNIST}} & \multicolumn{1}{c}{\textbf{CNN+CIFAR-10}} & \multicolumn{1}{c}{\textbf{ResNet18+CIFAR-10}} & \multicolumn{1}{c}{\textbf{ResNet18+Fruits}} \\
\midrule
1  & 2.2696 & 1.4554 & 0.7428 & 1.2410 & 1.1479 \\
2  & 1.3309 & 1.1270 & 1.1401 & 1.0079 & 0.9685 \\
3  & 2.4697 & 1.5957 & 0.7262 & 0.8968 & 0.7416 \\
4  & 2.1020 & 1.7977 & 0.6645 & 1.0381 & 0.8900 \\
5  & 2.4044 & 1.9396 & 0.9657 & 0.9546 & 0.7755 \\
6  & 2.6173 & 1.8368 & 0.8201 & 1.0064 & 0.8260 \\
7  & 3.1527 & 1.7818 & 0.9014 & 0.9752 & 0.8893 \\
8  & 2.7789 & 1.6647 & 1.1736 & 0.9933 & 0.8807 \\
9  & 1.6579 & 1.4312 & 0.8501 & 1.2456 & 0.8843 \\
10 & 2.1925 & 1.5532 & 1.0184 & 1.3401 & 1.0503 \\
\bottomrule
\end{tabular}
}
\end{table}

\subsection{Experiment results}

The first experiment, as presented in Table \ref{tab1}, aims to validate the widespread occurrence of singular value stratification in classification tasks. It encompasses multiple datasets and various classifier architectures that achieved high accuracy after training. For each dataset, we computed the Jacobian matrix of the data and performed singular value decomposition (SVD) to observe whether stratification of the singular values exists. For each class, we statistically analyzed 50 samples and calculated the Coefficient of Variation (CV) of the singular values for each sample. CV here is defined as:
$$\text{CV}=\frac{\sigma^2}{\mu^2}$$
where $\sigma$ is the standard deviation of the singular values and $\mu$ is their mean. We computed the mean CV for the 50 samples, and the results are shown in the Table.

Across different datasets and models, the CV of singular values consistently remains around or above 1. A CV close to 1 indicates that the deviation of each singular value from the mean singular value is approximately onefold. For example, in a sample where this CV is 1.2, the singular values are: $[30.03,8.63,7.48,5.31,4.13,3.09,2.95,2.69,1.80]$, which exhibit clearly strong stratification.

\begin{figure*}[!t]
    \centering
    \captionsetup[subfigure]{
                        textfont = rm,
                        labelfont = {rm, bf},
                        size = scriptsize,
                        }
    \subfloat[][class avocado]{\includegraphics[width=0.24\textwidth]{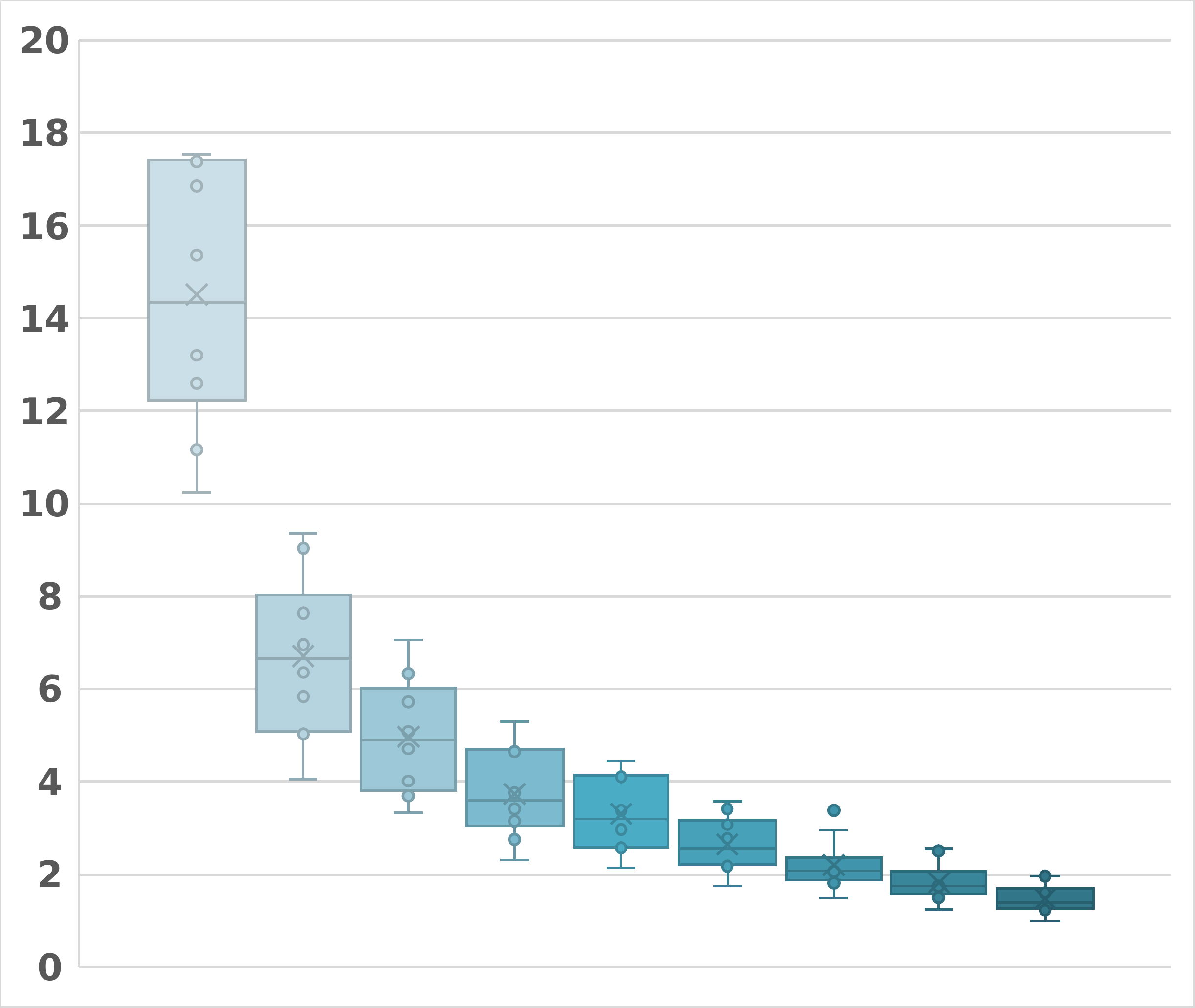}\label{fig2_first_case1}}
    \hfill
    \subfloat[][class watermelon]{\includegraphics[width=0.24\textwidth]{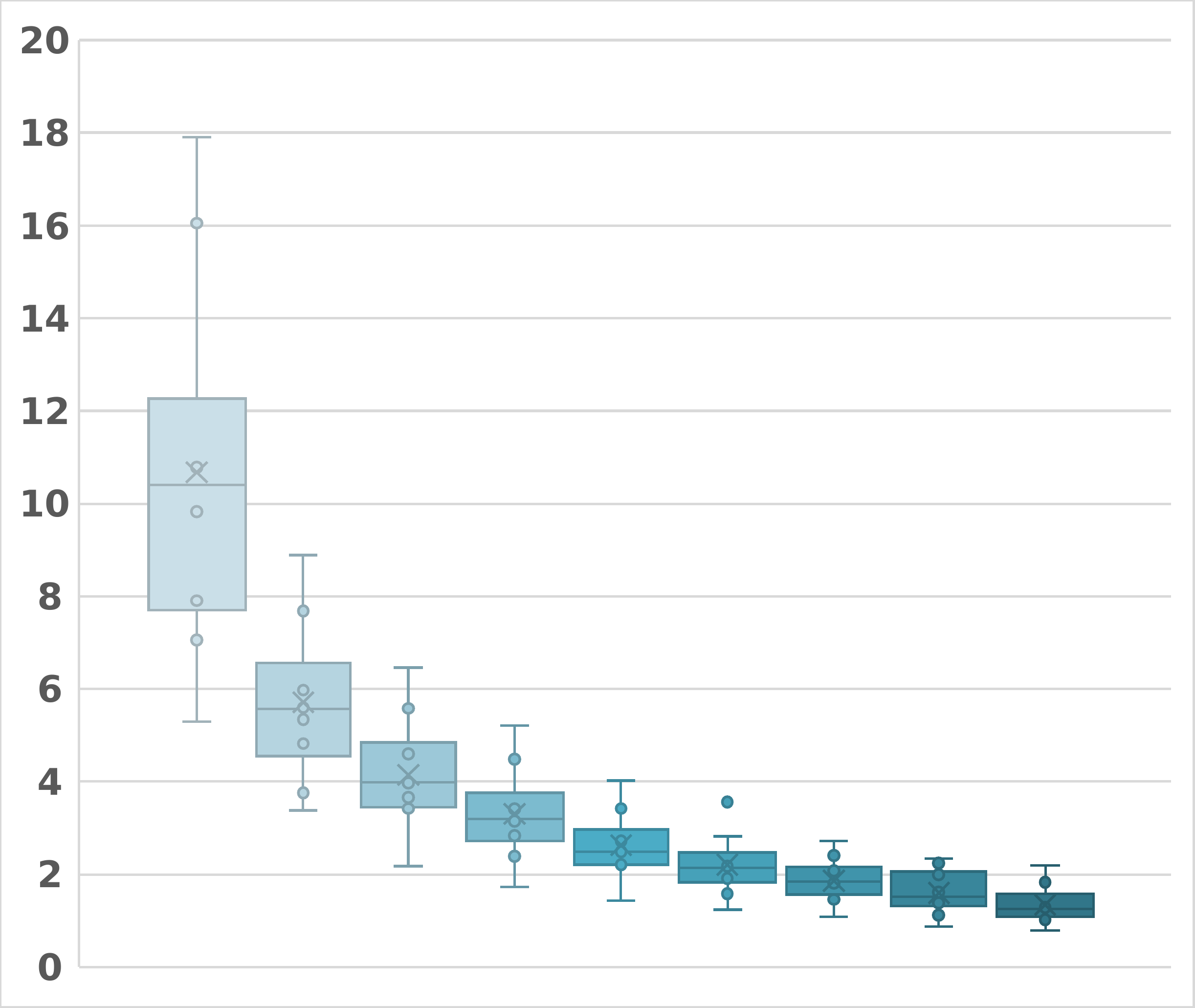}\label{fig2_second_case1}}
    \hfill
    \subfloat[][class noise image]{\includegraphics[width=0.24\textwidth]{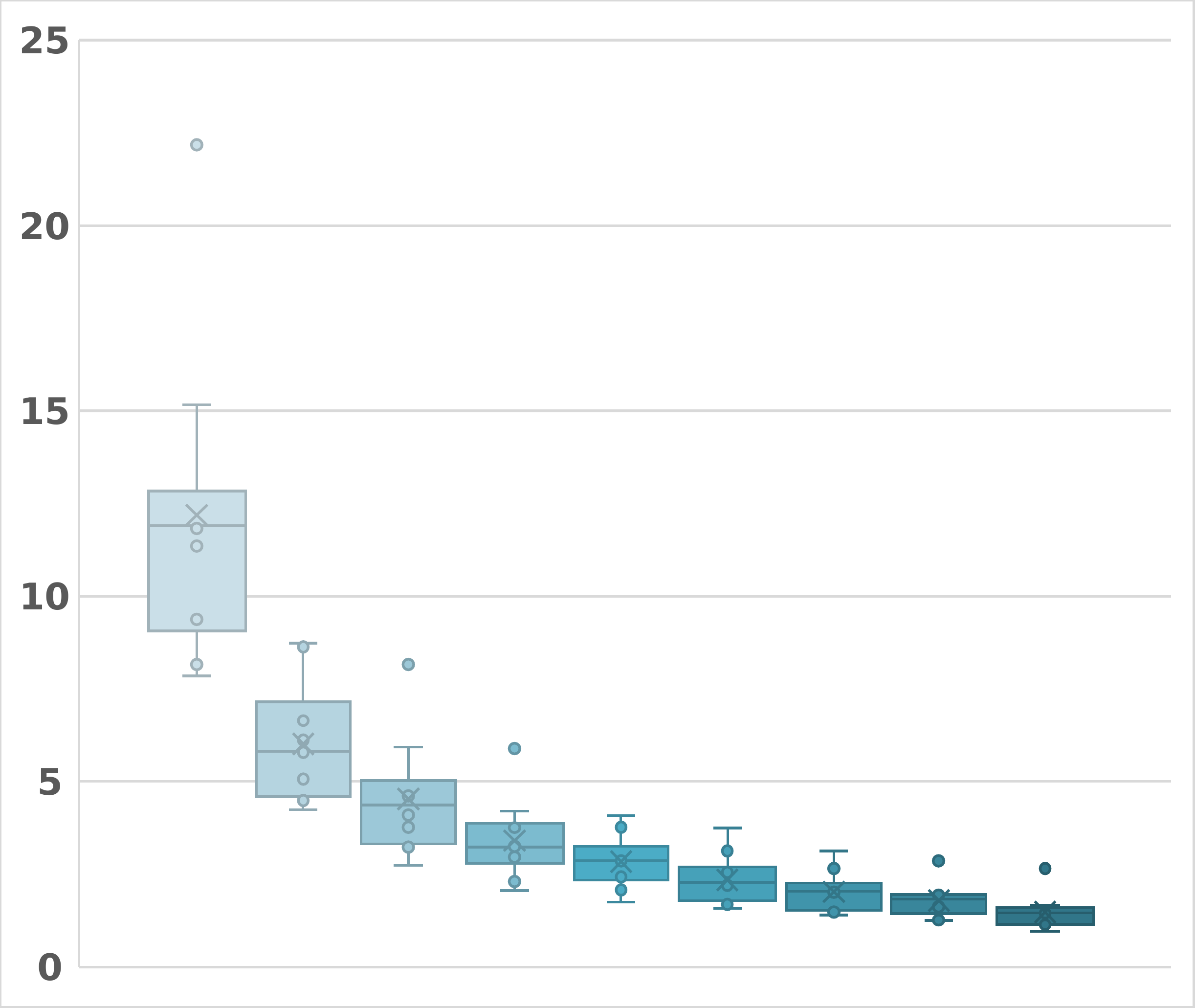}\label{fig2_first_case2}}
    \hfill
    \subfloat[][class noise]{\includegraphics[width=0.24\textwidth]{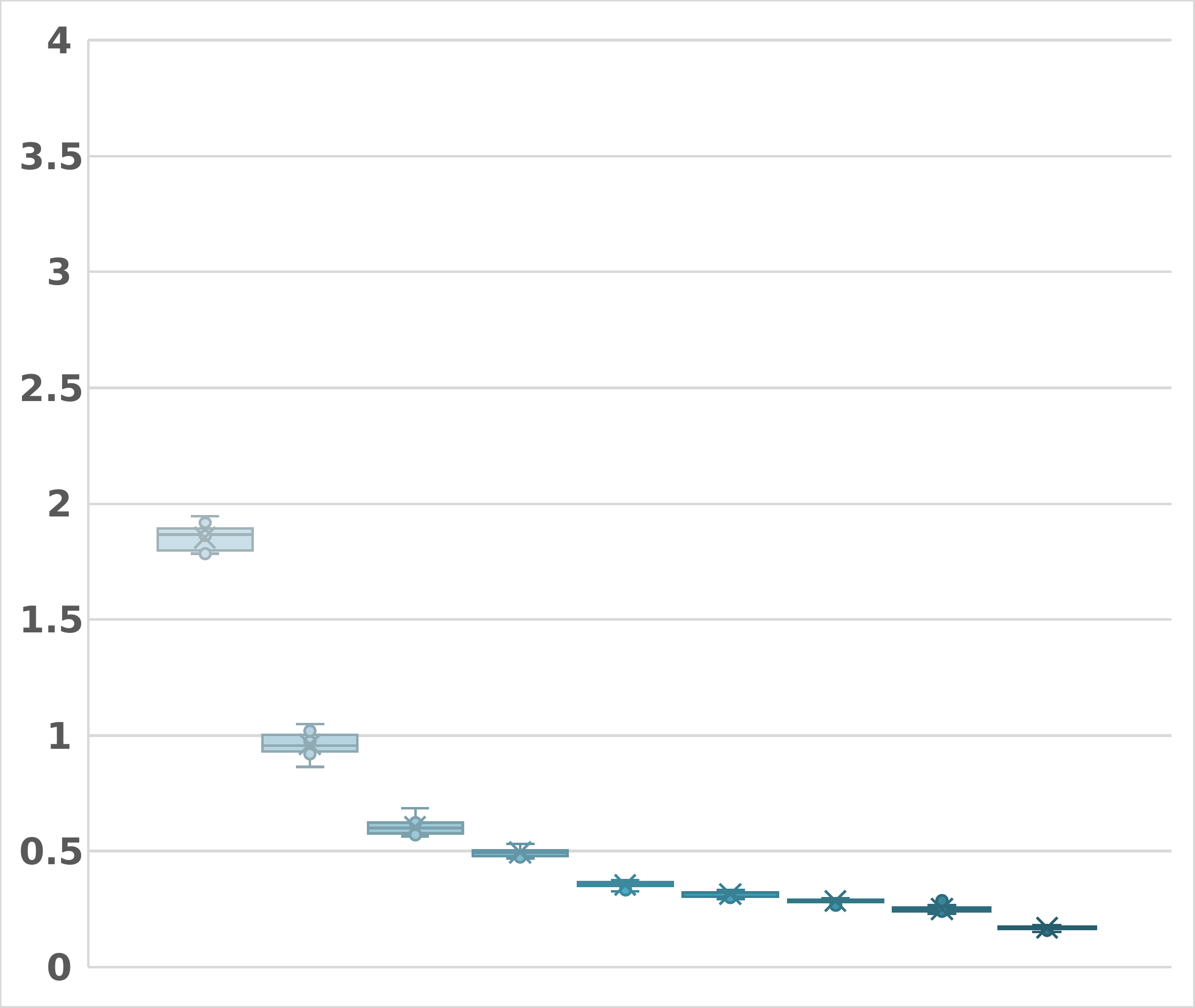}\label{fig2_second_case2}}
    \hfill

    \caption{Box plot of class-specific singular value decomposition(SVD) of classification results from a pre-trained ResNet-18 model(10 samples). Each dot corresponds to a singular value of a particular sample, and the $\times$ corresponds to the mean value of each group of singular values.}
    \label{fig_2}
\end{figure*}

The second experiment in this study investigates singular value stratification using a ResNet-18 classifier trained on a Fruit dataset. The training and testing setup is mentioned above. The results of SVD are illustrated in Figure \ref{fig_2} as box plots, which visually summarize the singular value decomposition using the median as a central line within the box, the quartiles as the box boundaries, and the mean value marked by an "×" symbol.

The singular value decompositions for the other fruit classes exhibit trends similar to that of avocado and are therefore omitted for brevity. Notably, all classes except the noise class exhibit a certain degree of stratification, suggesting that the model retains a degree of generalization for natural images, whereas the stratification pattern is much less apparent for purely random inputs.

\begin{figure}[!t]
    \centering
    \captionsetup[subfigure]{
                    textfont = rm,
                    labelfont = rm,
                    size = normalsize,
                    }
    \includegraphics[width=0.96\textwidth]{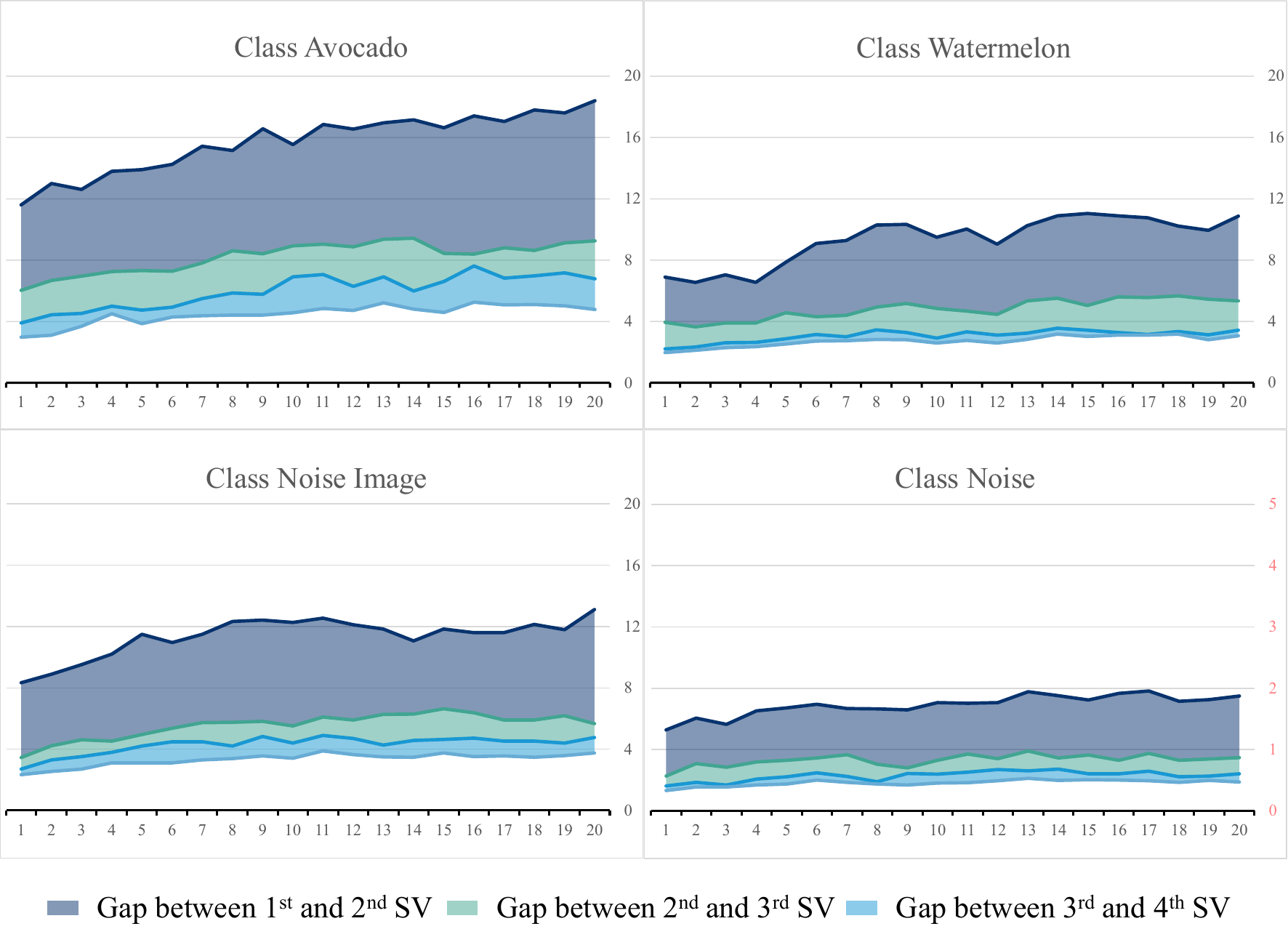}

    \caption{The first 4 singular values' trajectory as the ResNet-18 training process proceeds on the Fruit dataset.}
    \label{fig_3}
\end{figure}

\begin{figure}[!t]
    \hspace*{-0.9cm}
    \includegraphics[width=1\textwidth]{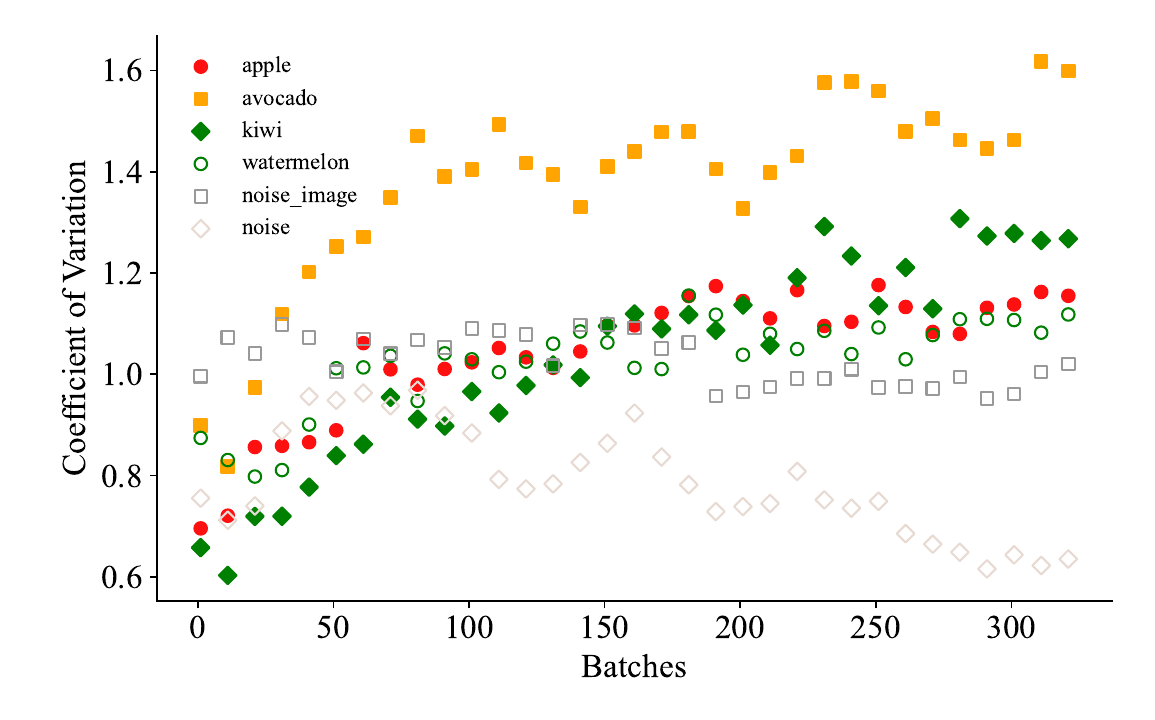}
    \caption{The evolution of the coefficient of variation (CV) of the SVD for specific images during the training process of a ResNet-18 classifier on the Fruit dataset of the first 350 batches in the first epoch. }
    \label{fig_4}
\end{figure}

We track the singular value along the training process, and the result is illustrated in Figure \ref{fig_3}, in which SV is the abbreviation of singular value. It can be observed that the structural differences occur in the early stage of the training process and become more significant as the training proceeds. Therefore, we also examine how the coefficient of variation for specific images evolves throughout the first training epoch, and the results are presented in Figure \ref{fig_4}. It can be observed that the training samples exhibit significantly higher CV values compared to noise images and noise samples, suggesting distinct structural differences in the learned representations. 

All experimental results indicate that during the training process of typical deep neural networks, the singular values of the Jacobian matrix of the model gradually exhibit a stratification phenomenon as training progresses. Based on our theory, we have strong reason to speculate that during training, sample points gradually converge to low-dimensional attractor manifolds. Moreover, for meaningful images in the vicinity of the samples that were not involved in the training, the singular values also show stratification. This is an indication of the model's generalization ability. We believe that extending attractor theory from recurrent neural networks to general neural networks is highly promising and meaningful.

\vspace{12pt} 

\vspace{12pt} 

\vspace{12pt} 

\section{Conclusion and Future Work}\label{sec6}
This study introduces a novel approach for investigating continuous attractors, providing a unified explanation for various phenomena and findings in prior attractor research. The proposed method, based on the eigenvalues or singular values of the local Jacobian matrix of a mapping, serves as a tool for determining the dimensionality of continuous attractors. By integrating our theoretical framework with existing studies, we offer an interpretation of approximate continuous attractor phenomena observed in neural networks. Furthermore, we examine the prevalence of eigenvalue and singular value stratification—phenomena closely associated with approximate continuous attractors—within neural network classification tasks.

However, this work currently lacks a rigorous mathematical derivation extending eigenvalue stratification to singular value stratification. Additionally, the existence of continuous attractors in tasks beyond classification remains an open question. Building upon the mathematical foundations established here, future research may develop a general theory of continuous attractors in neural networks, offering new insights into neural network interpretability.

\end{document}